# Designing Trustworthy AI:
# A Human-Machine Teaming Framework to Guide Development

**Carol J. Smith**

Carnegie Mellon University's Software Engineering Institute, Pittsburgh, PA
cjsmith@sei.cmu.edu

**Abstract**

Artificial intelligence (AI) holds great promise to empower us with knowledge and augment our effectiveness. We can—and must—ensure that we keep humans safe and in control, particularly with regard to government and public sector applications that affect broad populations. How can AI development teams harness the power of AI systems and design them to be valuable to humans?

Diverse teams are needed to build trustworthy artificial intelligent systems, and those teams need to coalesce around a shared set of ethics. There are many discussions in the AI field about ethics and trust, but there are few frameworks available for people to use as guidance when creating these systems. The Human-Machine Teaming (HMT) Framework for Designing Ethical AI Experiences described in this paper, when used with a set of technical ethics, will guide AI development teams to create AI systems that are accountable, de-risked, respectful, secure, honest, and usable.

To support the team's efforts, activities to understand people's needs and concerns will be introduced along with the themes to support the team's efforts. For example, usability testing can help determine if the audience understands how the AI system works and complies with the HMT Framework. The HMT Framework is based on reviews of existing ethical codes and best practices in human-computer interaction and software development.

Human-machine teams are strongest when human users can trust AI systems to behave as expected, safely, securely, and understandably. Using the HMT Framework to design trustworthy AI systems will provide support to teams in identifying potential issues ahead of time and making great experiences for humans.

## Introduction

Diverse teams are needed to build trustworthy artificial intelligent (AI) systems, and those teams need to coalesce around a shared set of ethics. The experience of working with people who are significantly different from us can be challenging, and studies have shown (Rock, 2019) diversity is worth the effort, as it increases our capacity for innovation and creative thinking. The shared set of ethics acts as a central point from which commonalities can be drawn, strengthening the team and their work.

Having a set of shared professional ethics helps smooth differences, to create an inclusive environment for the entire organization. In addition, a set of technology ethics, focused on the work being done, is equally important. The author recommends initially adopting an existing set of ethics from a reputable organization that is focused on computing and AI. The Montréal Declaration for a Responsible Development of Artificial Intelligence 2018: https://www.montrealdeclaration-responsibleai.com (Université de Montréal, 2018; Fjeld, et al, 2019; Hagendorff, 2019), is a respected starting place. As the organization matures, efforts can be made to adapt and customize a set of ethics to the needs of the team and the work.

An ethics code without usable guidance and/or training is less likely to be implemented as intended. The HMT Framework is to be used in conjunction with the team's technology ethics code, as a tool to support the inspection process and gauge if the process and the resulting system is indeed trustable. The HMT Framework is meant to bridge the gap between ethical stances such as "do no harm," and the reality of applying that stance to making new technology for humans to interact with.

To be effective in this work, teams must be diverse with regard to gender, race, education, thinking process, disability status, and more, as well as their skill set and problem framing approach. This includes a multi-disciplinary team mixing machine learning experts, programmers, system architects, product managers, and those who work as curiosity experts (CE), focused on understanding the situation, the constraints, and abilities of the people who will use the system and how it would be used. CEs comprise a broad range of job titles and skills including human-computer interaction and human-machine interaction professionals, cognitive psychologists, digital anthropologists, and user experience (UX) researchers. People without these job titles and skills should also participate in (and potentially lead) CE-related activities described in this paper. These CE activities will enable the team to uncover potential issues before they arise. The goal with bringing these diverse individuals together is to reduce bias in the system and to account for a broad set of unintended consequences. "Building a company or team that actively breaks down systemic prejudice for its own sake translates to building products that avoid harming large swaths of the population," stated UX expert Dan Brown in his blog post "UX in the Age of Abusability" (https://green-onions.com/ux-in-the-age-of-abusability-797cd01f6b13).

Doing this work requires deep conversations and agreement within the team and across the organization about issues as they come up. While this work is time consuming, the serious and contentious discussions that will emanate are precious with regard to aligning the team prior to facing a difficult situation.

The HMT Framework will help drive these conversations towards a clear understanding of what the expectations are in specific situations and help the team to create mitigation plans for how they will respond. The HMT Framework can be used during the initial creation of the AI system and once it is in production. AI is still evolving, so this is a first step towards helping teams deal with the complexity inherent in these systems. As more is learned the HMT Framework can be updated appropriately.

There are four themes in the HMT Framework for Designing Ethical AI Experiences and they will be discussed in detail in this paper. AI systems need to be created to be:
1) Accountable to humans
2) Cognizant of speculative risks and benefits
3) Respectful and secure
4) Honest and usable

## Accountable to humans

AI systems must be built in ways that ensure that humans are always in ultimate control and responsible for all that the AI system will do. This is particularly significant with regard to decisions that affect a person's life, quality of life, health, or reputation. All decisions and outcomes must remain the designated responsibility of humans. This is both to ensure that the decision is made carefully, but also to maintain the role of AI systems in supporting humans.

Depending on the system the diverse team is creating, this may not be an obvious piece of guidance and this is precisely where the discussions need to begin. What is a decision or outcome in this situation? For example, if the system will prioritize potential outcomes, how might you show outcomes that were not prioritized, but are common? How might the system show outcomes that were not prioritized because they are rare (but might be appropriate for this situation)? What if the conditions change: how will the effect on the potential outcomes be shown?

How will the people making the system share responsibility for the system once it is in use? For example, what is the responsibility of the CEs and designers vs. that of the developers and finally the people deploying the product? How about the people making decisions based on the information the AI system provided? Individuals should be aware of their responsibility from the beginning; as the product matures, the responsibility may change. The goal of designating specific responsibility is not to find someone to blame, but rather to maintain human control and increase personal investment in the product. Individuals should be protected by the organizations they work for to some extent; but where work is attributable to individuals, those individuals should be held responsible.

This does not mean that the lowest-paid person should be saddled with this responsibility, or that only the highest-paid person is responsible. The idea is to spread responsibility, so that the influential people on the team who are making decisions know that they are responsible for the system's performance and how it communicates information to humans. When there are complex organizations working together or when an organization is selling their AI system to customers, the lines of responsibility can be confusing. This is another opportunity to have serious discussions before issues arise that need to be managed.

As was already mentioned, final decisions that affect a person's life, quality of life, health, or reputation, must be made by a human being. Examples include judicial decisions regarding incarceration; significant financial decisions, such as mortgage interest rates; medical treatment selection for life-threatening diseases; and hiring and firing in the workplace. Significant decisions that are made by an AI system must be appealable to a human. This is to ensure that humans remain in control and that context is always being considered. The AI system must always be able to be overridden or have decisions reversed by designated people. This helps ensure that context is being considered and that humans remain in control.

Making an AI system is currently a time consuming and difficult task, with many iterations. Team members may claim that the system is making decisions based on unknowable algorithms, that it is a black box. This type of erroneous talk is dangerous and irresponsible. If the AI system is unknowable, it should be turned off. The team is always responsible for what the AI system does and must retain control of it at all times. The system needs to be monitored and when unexpected results surface, it should be even more carefully monitored. Ideally, these will be revolutionary ways the AI system has chosen to frame information. In some cases, the AI system may organize information incorrectly, resulting in unhelpful and potentially harmful results. How that is conveyed to the people using it will determine how much the system is trusted. The CEs can conduct formative activities such as Wizard of Oz studies (Wizard of Oz, 2005), cognitive walkthroughs and concept studies, and if the system is live they can observe people using it, to help identify ways to improve these situations. Risks need to be controlled and the AI system needs to provide context to the people monitoring and using the system so that they can see relevant information such as data provenance and confidence levels to base decisions on.

## Cognizant of speculative risks and benefits

Risks to human's personal information and decisions that affect their life, quality of life, health, or reputation (Université de Montréal, 2018), need to be anticipated and then evaluated long before humans are using or affected by the AI system. As previously mentioned, the team must be diverse with regard to gender, race, education, thinking process, disability status and more as well as their skill set and problem-framing approach. A cross-functional and diverse team will uncover a broader set of issues than one with primarily shared experiences. Including the people who will be using the system in this work will help to ensure that a broad set of scenarios are considered.

The diverse team needs to make time to identify the full range of harmful and malicious use of the AI system. This can be done in a variety of ways, the ideal one being a workshop or a series of workshops where information about the system's use is shared and then various potential issues are raised. This work does not need to be exhaustive or even intensive if the team feels a set of significant scenarios effectively covers the potential issues. The key is to make sure that a broad set of harmful and malicious use situations are identified and then evaluated and that mitigation techniques are established and agreed upon.

The team will use this time to identify and evaluate blind spots in the data – areas that perhaps the AI system does not know about, and may be confusing. For example, there are many homonyms in the English language that could be difficult for the system to understand; there may be acronyms used that mean different things in different contexts; and people often share the same names.

Unintended consequences can be numerous depending on the system, and at least the worst-case situations need to be identified. Casey Fiesler, in her 2018 blog "Black Mirror, Light Mirror: Teaching Technology Ethics Through Speculation," states that "encouraging creative speculation is an important part of teaching ethical thinking in the context of technology" (https://howwegettonext.com/the-black-mirror-writers-room-teaching-technology-ethics-through-speculation-f1a9e2deccf4). Aaron Lewis recommends conducting CE activities such as Black Mirror Brainstorms (https://twitter.com/aaronzlewis/status/1063544871472914432) to identify these important areas of concern. An additional CE activity that can be useful is evaluating the product's abusability with the intent of preventing it as Dan Brown recommends in his blog post "UX in the Age of Abusability" (https://greenonions.com/ux-in-the-age-of-abusability-797cd01f6b13). As Brown describes, we can extend existing tools and methods to consider abusability. These activities enable teams to think through some of the worst potential scenarios and help to identify ways to avoid those situations as a team.

Situations that are not preventable (such as the system being abused or trained maliciously) are identified early through these activities, and the team can determine what to do and how to react. The team's reactions will also need to be explored for consequences, and the team will need to create mitigation plans. For example, if the solution to a particularly negative or dangerous situation is to shut down the AI system, who has access and ability to do that? What are the consequences of shutting the AI system down?

There will need to be a way to communicate the situation to stakeholders as well as to the people relying on the system. How will they become aware of a situation? What resources are available to them as a backup? If the team is able to revert to a previous model, what knowledge is missing? How will this impact work? Planning those communications enables the team to use rationale and clear thinking in the response, rather than an in-the-moment response that has the potential to further exacerbate an already tense situation.

Part of the mitigation discussions should also include the ways that people using the system can report issues. Can they get clarification of the system status? Who is able to report issues and to whom? Once an issue is reported how will it be investigated? Who is responsible for escalation and who is the issue escalated to? How will you manage the AI system during non-business hours in a way that prioritizes protecting people without creating an unhealthy workplace? This work to determine what the team stands for, what the AI system will and will not do, and how to protect humans is all guided by the agreed-upon technology ethics.

## Respectful and secure

To gain trust of humans, AI systems need to be respectful and secure. This work starts with a team that values humanity, ethics, equity, fairness, accessibility, diversity, and inclusion, and transfers those values into their work. The diverse teams we bring together should include diversity in the people making and curating the content for initial training, the people involved in the training of the system, and the people who will monitor and manage the production system. The meaning of diversity will vary between these groups: in some cases, the focus will be on educational differences or work experience, or other differentiators beyond the obvious gender, race, cultural and disability status, and more.

Most ethics codes cover privacy and data rights, necessary for gaining trust and for a secure system. Ensuring that the people using the system feel their information is safe and not mandating that they provide more information than necessary is an important first step. As you are considering adding new features, creating user profiles, and making forms and surveys, evaluate what is the bare minimum information needed to do what is requested. Products accessible in the EU are required to meet the General Data Protection

Regulation (GDPR - https://eugdpr.org/) for data protection and privacy, and GDPR is a good tool for guidance in any market.

AI systems need to be accountable and transparent to the humans who made them, monitor them, and use them. People should be able to verify what the AI system is doing, and why, in a timely manner. As was mentioned in the accountability theme, the system cannot be considered unknowable or thought of as a black box. Retaining control of the AI system includes monitoring it for usage, outcomes, accuracy, confidence, and overall analysis. If people using the system suspect an issue, they should be able to do some research themselves (looking at confidence and accuracy if unexpected results are provided, for example) and if necessary, report issues. It should be clear what security methods are being used and how they protect the data, the AI system, and the people using the system. These are key parts of making people feel safe and actually keeping them safe.

Last, the system must be built in a robust, valid, and reliable way. If it is, the people who take care of it and those who use it can rely on it, and the people intending to do harm have less ability to do so. The AI system needs to be technically safe and built in a professionally acceptable way, to current standards, so that individuals new to working on the system are able to relatively quickly understand how it was built.

All of these efforts can be evaluated though CE activities such as interviews with individuals creating, maintaining, and using the system to understand concerns and issues. The CEs can also review metrics to look for patterns or anomalies. Finally, the team can use standard test procedures to determine the efficiency of the system and other metrics.

## Honest and usable

Last, but not least, AI systems must be designed to be honest and usable. An honest and usable system values transparency with the goal of engendering trust of everyone interacting with it. This includes explaining the AI system and its limitations in language that the audience understands. For example, a new user to the system should be able to ascertain what it does and how it works. The limitations should be provided in plain language that is easily understood.

For example, a facial recognition system may be biased due to being trained on primarily white faces and may not recognize darker skin. Similarly, a voice-to-text system trained on American English may recognize neither accents from outside the U.S. nor other forms of English. By being forthright with admissions of bias and weakness in the system, people can ascertain for themselves if it will be useful to them. A diverse team can reduce the chances that the team creates solutions that reflect their own biases--such as computer vision systems that only recognize white faces as "people." Just like the humans creating it, no system is perfect. It will have limitations, biases and other imperfections that should be shared clearly with a sense of humility to the people interacting with the system.

The system should align with basic usability heuristics (Nielsen, 1994), particularly with regard to explicitly stating status and interaction points so that people using the system are not left wondering what it is doing. People interacting with the system should be able to easily discern when the AI system is taking action and/or making decisions. As was mentioned previously, significant decisions that are made by an AI system must be appealable, and access must be provided to transparent justification for those outcomes.

An honest AI system will provide humans with visibility to itself so that humans can easily discern when they are interacting with an AI system rather than another human. Whether by voice or by typing/texting, the AI should make its presence known and identify itself accurately to be trustworthy. Tricking a human into thinking they are communicating with another human, such as a Turing test (Oppy, Dowe, 2019) to determine sentience, is fine in a lab with full consent of the participants. But that is not an acceptable way for an AI system to interact with humans day-to-day.

The AI system needs to present and explain data sources, their provenance, and the training methods in both technical and plain language. Humans are all inherently biased and per Buster Benson, without bias' providing "the ability to act fast in the face of uncertainty, we surely would have perished as a species long ago" (https://medium.com/better-humans/cognitive-bias-cheat-sheet-55a472476b18). However, we now know to push to overcome these biases to benefit from the combining of diverse teams. That same bias is always present in anything we create--including data sets, training sets, and the resulting models--unless we are extremely careful. Human nature is to skip the difficulty of pointing out the bias and explaining it and instead rely on the presence of "data" as the proof of a lack of bias. This is not good enough for a trustworthy system of any type and definitely not an AI system. All bias needs to be recognized and explained clearly in AI systems.

As the system evolves and incorporates new information, the status of updates needs to be conveyed to users. When possible, those updates should be scheduled so that people using the system can anticipate if it will affect their work and have the opportunity to schedule work around the updates. The AI system's evolution should include regular cycles of improvements to meet both human needs and technical standards. The monitoring systems, as has been mentioned, should be designed in straightforward ways that are easily interpretable for everyone who needs to use them.

To support the work of creating honest, usable designs, CEs can conduct activities such as concept tests and usability studies with the people using (or who will be using) the system. The studies might include tasks such as asking people to find information and its provenance in the system, to determine when the AI system was last updated, and to determine the system's status. Those and other activities will give the team guidance for future design and development improvements.

## Human-Machine Teaming Framework

Use the HMT Framework to guide development of accountable, de-risked, respectful, secure, honest and usable AI systems, with a diverse team aligned on shared ethics.

A form for your team to use as a checklist and to sign in agreement with the HMT Framework principles, is available online as a PDF: https://drive.google.com/open?id=1aI-oJb2henbufT5eZ2MxTrQfdWrplYFc2

**We are confident that we have designed our AI system so that:**

We ensured humans are always in control, able to monitor and control risk.
We designated responsibility to humans for all decisions and outcomes.
We explicitly defined responsibility and who shares responsibility.
We preserved human responsibility for final decisions that affect a person's life, quality of life, health, or reputation.
Significant decisions made by the AI system are:

- Appealable
- Able to be overridden
- Reversable

**We identified the full range of risks and benefits:**

- Harmful, malicious use
- Good, beneficial use
- Blind spots and unintended consequences

**We have created plans:**

- Communication plan(s) for misuse/abuse of AI system
- Mitigation plans for misuse/abuse of AI system

**The AI system is respectful and secure:**

We integrated our values of humanity, ethics, equity, fairness, accessibility, diversity and inclusion.
We respected privacy and data rights - only necessary data is collected, not more.
We provided understandable security methods.
The AI system is robust, valid and reliable.

**We value transparency with the goal of engendering trust:**

The purpose and limitations of the AI system are explained in plain language.
Data sources and training methods have unambiguous sources and are verifiable.
Confidence and context are presented for humans to base decisions on.
We provided transparent justification for outcomes.
The AI system includes straightforward, interpretable, monitoring systems.

**The AI system explicitly states its identity, is honest and usable:**

Humans can easily discern when they are interacting with the AI system vs. a human.
Humans can easily discern when and why the AI system is taking action and/or making decisions.
Improvements will be made regularly to meet human needs and technical standards.

## Closing

Trustworthy AI systems require diverse teams that coalesce around a shared set of technology ethics. The shared set of ethics acts as a central point from which commonalities can be drawn, strengthening the team and their work. Conducting CE activities to understand people's needs and concerns for the system will help to identify the risks and benefits of the system. Using the Human-Machine Teaming Framework for Designing Ethical AI Experiences alongside technology ethics will guide AI development teams in creating AI systems that are accountable, de-risked, respectful, secure, honest, and usable.

Successful organizations will have brought diverse teams together resulting in reduced bias and will empower teams to have deep conversations to align prior to facing a difficult situation. The result is clear expectations and mitigation plans for responding in constructive ways that protect people. AI is still evolving, and this first step towards helping teams deal with the complexity inherent in these systems will be built upon as the work on AI systems progresses.

## References

ACM. 2018. ACM Code of Ethics and Professional Conduct. (22 June, 2018). Retrieved September 13, 2019 from: http://www.acm.org/code-of-ethics

Fjeld, J.; Hilligoss, H.; Achten, N.; Daniel, M.L.; Kagay, S.; Feldman, J. 2019. Principled Artificial Intelligence: Mapping Consensus and Divergence in Ethical and Rights-Based Approaches. Project. Berkman Klein Center for Internet and Society at Harvard

University. Boston, MA. Data Visualization retrieved September 13, 2019 from: https://ai-hr.cyber.harvard.edu/primp-viz.html

Hagendorff, T. 2019. The Ethics of AI Ethics: An Evaluation of Guidelines. University of Tuebingen. International Center for Ethics in the Sciences and Humanities. Retrieved September 16, 2019 from: https://arxiv.org/abs/1903.03425

Nielsen, J. (1994). Enhancing the explanatory power of usability heuristics. ACM CHI'94 Conference (pp. 152-158). Boston, MA: Association for Computing Machinery.

Oppy, G.; Dowe, D. 2019. The Turing Test. Retrieved from The Stanford Encyclopedia of Philosophy (Spring 2019 Edition), Edward N. Zalta (ed.): https://plato.stanford.edu/archives/spr2019/entries/turing-test/

Rock, D.; Grant, H. 2019. Why Diverse Teams Are Smarter. (4 November, 2019). Retrieved September 13, 2019 from: https://hbr.org/2016/11/why-diverse-teams-are-smarter

Université de Montréal. 2018. Montréal Declaration for a responsible development of artificial intelligence. 2018. Retrieved September 13, 2019 from: https://www.montrealdeclaration-responsibleai.com/the-declaration

Usability Body of Knowledge. 2012. Wizard of Oz. Retrieved from Usability Body of Knowledge, UXPA: https://www.usabilitybok.org/wizard-of-oz

# Acknowledgements

Copyright 2019 Carnegie Mellon University.

This material is based upon work funded and supported by the Department of Defense under Contract No. FA8702-15-D-0002 with Carnegie Mellon University for the operation of the Software Engineering Institute, a federally funded research and development center.

References herein to any specific commercial product, process, or service by trade name, trade mark, manufacturer, or otherwise, does not necessarily constitute or imply its endorsement, recommendation, or favoring by Carnegie Mellon University or its Software Engineering Institute.